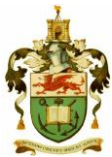 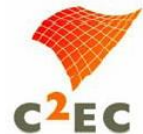

# Particle Swarm Optimization: Development of a General-Purpose Optimizer

M. S. Innocente[†] and J. Sienz[†]

[†]University of Wales Swansea, Centre for Polymer Processing Simulation and Design, C²EC
Research Centre, Swansea, SA2 8PP, Wales-UK.
mauroinnocente@yahoo.com.ar    J.Sienz@swansea.ac.uk

### Abstract

For problems where the quality of any solution can be quantified in a numerical value, optimization is the process of finding the permitted combination of variables in the problem that optimizes that value. Traditional methods present a very restrictive range of applications, mainly limited by the features of the function to be optimized and of the constraint functions. In contrast, evolutionary algorithms present almost no restriction to the features of these functions, although the most appropriate constraint-handling technique is still an open question. The particle swarm optimization (PSO) method is sometimes viewed as another evolutionary algorithm because of their many similarities, despite not being inspired by the same metaphor. Namely, they evolve a population of individuals taking into consideration previous experiences and using stochastic operators to introduce new responses. The advantages of evolutionary algorithms with respect to traditional methods have been greatly discussed in the literature for decades. While all such advantages are valid when comparing the PSO paradigm to traditional methods, its main advantages with respect to evolutionary algorithms consist of its noticeably lower computational cost and easier implementation. In fact, the plain version can be programmed in a few lines of code, involving no operator design and few parameters to be tuned. This paper deals with three important aspects of the method: the influence of the parameters' tuning on the behaviour of the system; the design of stopping criteria so that the reliability of the solution found can be somehow estimated and computational cost can be saved; and the development of appropriate techniques to handle constraints, given that the original method is designed for unconstrained optimization problems.

### INTRODUCTION

Optimization is the process of seeking the combination of variables that leads to the best performance of the model, where "best" is measured according to a pre-defined criterion, usually subject to a set of constraints. Thus, setting different combinations of values of the "variables" allows trying different candidate solutions, the "constraints" limit the valid combinations, and the "optimality criterion" allows differentiating better from worse. Traditional optimization methods exhibit several weaknesses such as a number of requirements that either the function to be optimized or the constraint functions must comply with for the technique to be applicable, and their usual incapability of escaping local optima.

Evolutionary algorithms (EAs) comprise a number of techniques developed along the last few decades, which are inspired by evolution processes that natural organisms undergo to adapt to a dynamic environment in order to survive. Since these organisms adapt by seeking the best response to the challenge they are facing, they happen to perform complex optimization processes, which can be viewed as processes of fitness maximization. It is important to remark that, since they do not specifically intend to perform optimization but to adapt to the environment, it is frequently claimed that they are not "optimization" but "adaptation" methods. It turns out that such adaptation results in optimizing the fitness of the individuals. Although these methods typically require higher computational resources than traditional methods, they do not impose restrictions on the features of the function to be optimized or on the formulation of the constraints. Last but not least, they are not problem-specific but general-purpose methods, which require few adaptations or none to deal with different problems, as opposed to traditional methods.

On the one hand, EAs can be viewed as "modern heuristic techniques" because they are not developed in a deterministic fashion. That is to say that they are not designed to optimize a given problem but to perform some procedures which are not directly related to the optimization process. Optimization occurs, nevertheless, despite there not being clear, evident links between the implemented technique and the resulting optimization process[1]. On the other hand, EAs can also be viewed as "Artificial Intelligence (AI) techniques"[2], because their ability to optimize is an emergent property that is not specifically intended, and therefore not implemented in the code. EAs are not designed to optimize but to carry out some kind of artificial evolution performing biological-like evolution processes such as mutation, recombination, and selection, which results in the maximization of a fitness function that resembles biological evolution. Thus, the boundaries between the fields of optimization and AI become vague, and the optimization field becomes multidisciplinary, involving mathematics, computer science, engineering, genetics, and social psychology, to name a few.

---

[1] Detractors of modern heuristics argue that using them implies giving up on understanding the real problem.

[2] More precisely, "Artificial Life (AL) techniques".

Swarm intelligence (SI) is the branch of AI which is concerned with the study of the collective behaviour that emerges from decentralized and self-organized systems. It is the property of a system whose individual parts interact locally with one another and with their environment, inducing the emergence of coherent global patterns that the individual parts are not aware of. That is to say that their awareness is limited to the local interactions, without having a sense of purpose of the global emergent behaviour of the whole system. The key issue is the concept of emergence, which is still not completely understood. An emergent property is a feature of a swarm of simple entities as a whole, which does not exist at the individual level. The interactions among a number of entities might give birth to an emergent property, which is not possible to be inferred by analyzing an isolated individual. Likewise, when designing artificial entities that would display emergent properties, such properties cannot be implemented in a deterministic fashion. It is extremely difficult even to predict whether a property would emerge from certain kinds of interactions among certain kinds of entities (not to mention which property) because the interactions, which are executed based on purely local information, must generate a positive feed-back effect. Typically, a lower threshold for the number of entities involved is required for the feed-back to take place. However, the interactions may just cancel each other out.

There are so many similarities between the EAs and the particle swarm optimization (PSO) method that some researchers consider the latter as yet another EA, despite not being inspired by natural evolution. Nevertheless, it evolves a population of individuals by profiting from previous experiences and using stochastic operators to introduce new responses, very much like evolution. However, since the PSO method also adheres to the principles of SI articulated by Millonas (quoted in [1]), the method is also viewed as one of the most successful "SI-based problem-solving techniques".

The PSO paradigm was originally designed by social-psychologist James Kennedy and electrical-engineer Russell Eberhart, in 1995 [1]. Although the method was inspired by previous bird flock simulations, the latter were framed within the field of social psychology, under the sociocognitive view of mind[3]. Therefore, the paradigm is also closely related to other simulations of social processes, having strong roots in both artificial life (AL) and social psychology. From the optimization point of view, it is a global method capable of dealing with optimization problems whose solutions can be represented as points in an $n$-dimensional space. In its original version, the design variables must be real-valued, although binary versions of the method were developed (e.g. [2, 3]), and attempts to handle discrete problems were also carried out (e.g. [4, 5]).

The PSO approach and the EAs are population-based methods that rely on stochastic operators to introduce creativity. They are bottom-up approaches in the sense that the system's intelligent behaviour emerges in a higher level than the individuals', evolving intelligent solutions without using programmers' expertise on the subject matter. While this feature makes it difficult to understand the way optimization is actually performed, these algorithms show astonishing robustness in dealing with many kinds of complex problems that they were not specifically designed for[4], as opposed to traditional optimization methods. However, these robust, general-purpose optimizers have the disadvantage that their theoretical bases are extremely difficult to understand in a deterministic fashion, if not impossible. Although much theoretical work has been carried out, only problem-specific and partial conclusions have been achieved in such important matters as the convergence and the tuning of the algorithms' parameters. The truth is that their precise behaviour is not fully understood, what should be of no surprise considering that they are not designed in a fully deterministic fashion. Therefore, attempting to understand them in that line of thought appears to be rather contradictory.

Some of the most mathematically rigorous works with regards to the particles' trajectory and the system's convergence were carried out by French mathematician Maurice Clerc (e.g. [6, 7, 8]). It is fair to note, however, that he studied extremely simplified systems[5], and the conclusions were extrapolated to the full systems. Nevertheless, the so-claimed guaranteed convergence was empirically verified on a test suite of benchmark functions.

The present paper intends to introduce the PSO method; to discuss its main strengths and weaknesses; and then it focuses on one of its widest spread versions, which is called here the "basic particle swarm optimizer" (B-PSO). The influence of each parameter of the algorithm on the behaviour of the swarm is analyzed in terms of the evolution of the best conflict[6] found so far; of the average of the current conflicts of the particles; of its ability to, and speed of clustering; and of its robustness[7]. The behaviour of the system heavily depends on the tuning of the parameters of the particles' velocity updating equation. For instance, the particles might be made more "self-confident" by assigning them higher individuality than sociality, which results in greater reluctance to becoming a follower, which in turn results in higher exploration being carried out in detriment of the speed of clustering. Thus, some tunings taken from the literature are tried and some others are proposed, discussed, and tested on a suite of benchmark functions.

---

[3] That is, thinking and intelligence as social phenomena.

[4] The individuals are indeed unaware of the fact that they are optimizing because they are not programmed to do so.

[5] He studied the trajectory of a single particle, removing the random weights from the particle's velocity updating equation, and keeping the two best previous experiences stationary.

[6] The function to be optimized is called the conflict function due to the social-psychology metaphor that inspired the method: the particles fly over the space of beliefs, seeking the minimization

Although optimization is ideally in quest for the best solution possible, this is often not the case in real-world problems, where successive improvement is already a great success. Besides, the concept of "possible" is remarkably subjective: usually, an optimization process is stopped because no further significant[8] improvement is being achieved, or because a maximum permissible number of time-steps have been reached. Therefore, this paper is also concerned with the design of stopping criteria, so that the iterative search can be terminated if further improvement is believed to be either unlikely or negligible, or if an upper threshold of time-steps has been reached. This allows both saving computational costs and estimating the reliability of the solution found. Although the latter cannot be directly measured, the fact that the search carried out by a very robust optimizer is terminated because further improvement is unlikely implies that the abilities of such an optimizer have been fully profited. In turn, saving computational costs is of utmost importance, given that the greatest practical disadvantage of the PSOs with respect to traditional methods is their higher computational requirements.

Finally, it is fair to remark that the original algorithm is suitable for unconstrained problems only. "Penalization methods" to handle constrained optimization problems such as those typically used for EAs are, in principle, suitable. Several other techniques can be found in the literature, with no one clearly outperforming the others in every case. That is to say, the best constraint-handing technique appears to be problem-dependent. A few techniques are briefly discussed within this paper, but only the "preserving feasibility" one is implemented due to the fact that it is straightforward and it can be applied, in principle, to any kind of constraint. In contrast, the application of other techniques such as the "cut off" of the particles displacement to other than hyper-cube-like boundary constraints is not so straightforward, while the "penalization" methods require problem-specific fine-tuning[9]. However, the "preserving feasibility" technique is not suitable for equality constraints straightaway, and it is typically not efficient in dealing with inequality constraints that lead to feasible regions of the search-space that are small in size or disjoint. Research on the development of constraint-handing techniques to handle such problems is currently ongoing, and the best alternative remains an active research question.

**ORIGINS**

The PSO paradigm was originally developed by social-psychologist James Kennedy and electrical-engineer Russell Eberhart in 1995 [1]. Despite the fact that it is mainly used in practice for optimization purposes, some principles underlying simulations of sociocognitive phenomena were of great influence for its development. In fact, although the method was inspired by previous bird flock simulations, such simulations were framed within the field of social psychology.

In the same fashion as the artificial neural networks (ANNs) can be viewed either as models of the human brain or as general mapping devices, and the GAs can be viewed either as models of genetic evolution or as optimization algorithms, the PSO paradigm can be thought of either as a model of social behaviour (e.g. a model of the spread of features through a culture) or as a problem-solving technique[10].

**Some influential experiments from social psychology**

In 1936, Sherif (quoted in [3]) reported experiments demonstrating the convergence of people's perceptions. He placed subjects in a dark room with a stationary spot of light projected on a wall. When asked in isolation, the individuals tended to report that the spot had been moving[11], although the range of the movement reported varied from person to person. However, when they were asked to make the report in public, the reports tended to converge. In 1956, Asch (quoted in [3]) reported that when subjects in an experiment were faced with the dilemma of giving the obvious true answer versus agreeing with the group, about a third of them chose to agree with the group in spite of knowing that the answer was plainly wrong. In 1965, Bandura (quoted in [3]) announced the discovery of the "no-trial learning", arguing that humans can learn a task without even

---

of the conflicts among the beliefs each particle holds by using the information gathered by their own and by other particles' experiences. They indirectly seek agreement by clustering in the space of beliefs, which is, broadly speaking, the result of all the particles imitating the most successful ones, thus becoming more similar to one another as the search goes by. The clustering is delayed by their own previous successful experiences, which the particles are reluctant to disregard, resulting in further exploration of different combinations of beliefs.

[7] Keep in mind that "robustness" refers to the optimizer's reluctance to getting trapped in suboptimal solutions within this work (not to be confused with "robust optimization").

[8] Again, the quantitative meaning of "significant" is subjective.

[9] Attempts to find general-purpose tunings for the parameters of "penalization methods" can be found in the literature.

[10] Notice that "optimization algorithm" and "problem-solving technique" can be viewed as synonyms, since any problem to be solved can be easily turned into an optimization problem (for instance, by defining an error function that is to be minimized).

[11] This is due to the "autokinetic effect", in the absence of any visual frame of reference (refer to [3], page 202).

trying it, by observing somebody else doing it with successful results.

Note that the tendency to seek agreement manifested in Sherif's experiment, the conformism observed in Asch's experiment, and Bandura's social learning, all support the belief that whenever people interact, they become more similar to one another. This is the key concept underlying some models of social behaviour such us "Axelrod's Culture Model" and the PSO paradigm.

Latané (quoted in [3]) suggested in his "social impact theory" that the influence of a group of people over an individual is a function of the strength, the immediacy, and the number of people in the group. The strength is just a kind of social persuasiveness, and the immediacy is inversely proportional to the distance. The influence increases—although the rate of increase decreases—with the number of individuals in the group.

**Some influential AL simulations**

It can be observed that some kinds of fish schools and bird flocks orderly wander in a rather majestic fashion. For instance, when a predator approaches a fish school, the fishes that first notice the threat change direction, and suddenly, they all change direction at what appears to be the same instant, so as to match their neighbours' new velocities. Some models of this behaviour have been proposed, suggesting that a single fish is attracted to a school, and that the attraction increases—although the rate of increase decreases—with the size of the school[12]. A few other simple rules prevent them from crashing into one another.

The behaviour of bird flocks is very similar, and many different models have been proposed. A well-known simulation of bird flocks was developed by Reynolds (quoted in [3]), who proposed three basic rules for each bird to follow:

1. Pull away before crashing into another bird.
2. Try to match the neighbours' velocities.
3. Try to move towards the centre of the flock.

Although the rules are entirely artificial, the simulation resulted in realistic flock-like behaviour. It is self-evident that biological animals try to avoid crashing, and that matching the neighbours' velocities is helpful in that regard. It is also reasonable to expect that social animals such as some kinds of fishes, birds, zebras, etc., would try to move towards the centre of the group because staying near the edge of the herd increases the chances of being hunted[13].

Another influential work was that of Heppner and Grenander (quoted in [1, 3]), who observed the critical issue that natural bird flocks do not have a leader. In other words, there is no central control! Heppner and Grenander implemented a simulation similar to that of Reynolds, but now the birds were also attracted to a roost, and an occasional random force was implemented seldom deflecting the birds' direction, resembling a gust of wind. The intensity of the attraction was programmed to increase with the decrease of the distance to the roost. The result was a realistic flock-like choreography.

**Origins of the PSO paradigm**

The paradigm was originated on the simulation of a simplified social milieu, where individuals were thought of as collision-proof birds. Thus, Kennedy et al. [1] modelled the graceful but unpredictable choreography of a bird flock in a 2-dimensional space, where collision was not an issue[14]. A first simulation was developed so that, at each time step, each artificial bird adopted its nearest neighbour's velocity, while a stochastic variable called "craziness" modified some randomly chosen velocities in order to prevent the simulation from settling on a unanimous, unchanging direction.

Heppner and Grenander's (quoted in [1, 3]) artificial birds were attracted to a roost (or to a food source), which led Kennedy et al. [1, 3] to think of optimization. However, those simulations profited from knowing the location of the "roost" in advance. In contrast, both real birds and the PSO algorithm search for, and eventually find, "food" without any prior knowledge regarding its location. Instead, they perform a parallel exploration of the environment, and profit from sharing the information gained by every individual.

While the emergent properties of the PSO paradigm result from local interactions among individuals within a population, Kennedy et al. [3] suggest that the behaviour of the individuals can be summarized in terms of three principles:

**1. Evaluate**: The organism evaluates the environment by evaluating the stimuli perceived by its sensors, in order to decide the proper reaction. Suppose, for instance, that each individual's mind is represented by an ANN, and each state of mind is defined by a set of weights. Every individual must be able to receive stimuli from the environment (inputs to the ANN) and make inferences (outputs from the ANN) at any time, thus evaluating the state of its mind. Note that an ANN can be represented by a particle of the PSO paradigm.

**2. Compare**: Once the stimuli are evaluated, it is not straightforward to tell good from bad. Experiments and theories in social psychology suggest that humans judge themselves by comparing to others (i.e. telling better from worse rather than good from bad). For instance, the strength in the social impact theory suggests that the persuasiveness of the individuals plays an important role in their influence over other individuals (successful individuals are more persuasive), and Bandura's no-trial learning suggests that humans can learn socially by imitating the behaviours of other successful individuals.

---

[12] Notice the similarities between the fish schools models and Latané's "social impact theory".

[13] It has been observed that this kind of social behaviour is more frequent in preys than in predators.

[14] Note that "collision" in the physical space is equivalent to "agreement" in the space of beliefs.

**3. Imitate**: Humans compare their own performances to those of others, and imitate only those individuals whose performance is superior or somehow desirable.[15]

While Kennedy et al. [3] arguably claim that nothing but these three processes occurs within the individual, it is merely noted here that these three processes are implemented within the PSO paradigm with remarkable success: the only sign of individual intelligence shown by the particles is a small memory. However, the PSO paradigm coupled with other paradigms can give birth to more intelligent artificial beings that can make inferences (for instance, by means of ANNs); **evaluate** the goodness of their own inferences; **compare** them to the goodness of other individuals' inferences, and perhaps also compare their inferences to those of other individuals who they have never even been in direct contact with, but whose performances are stored in the form of culture; and finally **imitate** the most successful inferences. In addition, some individual learning can be incorporated by means of a local search. Even further, all the individuals—and perhaps the culture—can be also subjected to some kind of biological-like evolution.

To the knowledge of the authors of this article, no algorithm this sophisticated has been implemented with proven success in dealing with engineering problems. However, it must be remarked that memetic algorithms consider biological-like evolution in the form of an EA coupled with individual learning in the form of a local search. Likewise, although the PSO approach already considers individual and social learning, embedding a local search would probably enhance the quality of the individual learning. Finally, it should be noted that many AL simulations consider several of this different phenomena acting together. For instance, Levy [9] studied the "Baldwin Effect"[16] by implementing a harsh artificial world inhabited by four different kinds of creatures, who had to learn which kinds of plants in the environment were comestible. Some beings had the ability to learn, some to evolve, some both abilities, whereas some had neither. In time, the only creatures left in the world were those with both abilities, which ended up knowing the information from birth. Since the AL-based problem-solving algorithms rely in finding the solutions themselves rather than implementing a deterministic and sequential procedure, it is reasonable to expect that artificial beings similar to these who adapted so well to a very harsh environment might be also able to cope with environments represented by extremely complex objective functions.

## BASIC PARTICLE SWARM OPTIMIZER

The bird-flock-like simulation of social behaviour was then generalized to *n*-dimensional collision-free search-spaces, where the equations that rule the trajectories of the particles were initially as follows:

$$v_{ij}^{(t)} = v_{ij}^{(t-1)} + iw \cdot U_{(0,1)} \cdot \left( pbest_{ij}^{(t-1)} - x_{ij}^{(t-1)} \right) + \\ + sw \cdot U_{(0,1)} \cdot \left( gbest_{j}^{(t-1)} - x_{ij}^{(t-1)} \right) \quad (1)$$

$$x_{ij}^{(t)} = x_{ij}^{(t-1)} + v_{ij}^{(t)} \quad (2)$$

Where:

- $x_{ij}^{(t)}$ : coordinate *j* of the position of particle *i* at time-step *t*
- $v_{ij}^{(t)}$ : component *j* of the velocity of particle *i* at time-step *t*
- $iw = sw = 2$ : individuality and sociality weights, kept constant and equal to 2
- $U_{(0,1)}$ : random number generated from a uniform distribution in the range $[0,1]$, resampled anew each time it is referenced[17]
- $pbest_{ij}^{(t-1)}$ : coordinate *j* of the best position found by particle *i* up to time-step $(t-1)$
- $gbest_{j}^{(t-1)}$ : coordinate *j* of the best position found by the swarm up to time-step $(t-1)$

As can be seen from equation **(1)**, a particle's velocity at a given time-step is equal to the velocity at the previous one altered by two components, one related to the particle's memory of its best previous experience, and the other related to the whole swarm's memory of its best previous experience. The individuality and sociality weights are set equal to one another and equal to 2, and random weights generated from the uniform distribution in the range $[0,1]$ introduce creativity into the system. Since the random weights are resampled anew for each time-step, for each particle, for each component, and for each term of equation **(1)**, the particles display odd, zigzagging trajectories that allow better exploration. In addition, the fact that the random weights are resampled anew for the individuality and the sociality terms, together with the individuality and sociality weights set equal to one another, makes each particle alternate between a more self-confident behaviour and a more conformist behaviour without any of them taking the lead for too long.

---

[15] Notice that the EAs also perform an **evaluation** of the individuals' performances, and the "survival of the fittest" requires the **comparison** between the individuals' performances, while breeding can be viewed as a kind of **imitation**, since it produces offspring that resemble their parents.

[16] The "Baldwin Effect" studies the chances that learning during a life-span affects the genetic evolution throughout generations.

[17] Beware that although the stochastic variable craziness was deleted, both the individual and social experiences are now affected by stochastic weights.

Given that the time-steps are increased in one unit at a time, the particles' positions are updated according to equation **(2)**, where $v_{ij}^{(t)} = \Delta x_{ij}^{(t)}$.

However, this algorithm presented a serious problem: the particles tended to diverge rather than cluster, so that the swarm appeared to perform a so-called "explosion". It was found that if the components of the particles' velocities were clamped, the explosion was controlled and the particles ended up clustering around a solution. An easy yet effective way of doing so is as follows:

**if** $\quad v_{ij}^{(t)} > v_{\max} \;\Rightarrow\; v_{ij}^{(t)} = v_{\max}$

**elseif** $\quad v_{ij}^{(t)} < -v_{\max} \;\Rightarrow\; v_{ij}^{(t)} = -v_{\max}$ (3)

The dynamics and reasons for the explosion to occur are still not completely understood, although they were found to be related to both the relative importance given to the second and third terms over the first one in equation **(1)**, and to the random weights. An example of the explosion for a 1-dimensional problem is shown in **Fig. 1** hereafter:

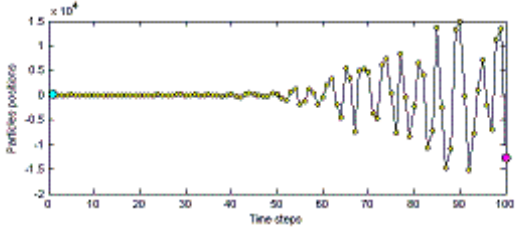

**Fig. 1**: Evolution of a single particle flying over a 1-dimensional search-space, where the two best values are fixed to zero, the particle is initially located at $x = 100$, its velocity is randomly initialized within the interval $[-1,1]$, no $v_{\max}$ is imposed, $iw = sw = 2$, and the function to be optimized is the Schaffer f6. The cyan and magenta dots are the particle's initial and final positions, respectively.

Clerc et al.[8] simplified the system in order to study the dynamics of the swarm from the bottom up (i.e. from a particle's point of view). Thus, the system was reduced to a single particle attracted towards two stationary best previous experiences, so that it was in reality attracted towards a fixed point that resulted from the weighted average of the two best stationary previous experiences. In addition, the random weights were simply removed from equation **(1)**. They proved that if $iw + sw < 4$, the particle exhibits a cyclic or quasi-cyclic behaviour. Even further, they found the particular values of $iw + sw$ for which the behaviour is cyclic. Conversely, there is no cyclic behaviour, and the particle diverges from $p$, if $iw + sw \geq 4$. The evolution of such a particle, whose velocity updating equation is given by:

$$v^{(t)} = v^{(t-1)} + (iw + sw) \cdot (p - x^{(t-1)})$$ (4)

where $iw + sw = 4$ and $p = 0$, is shown in **Fig. 2**. This divergence is called here the "deterministic explosion".

Clerc et al. [8] analytically developed a constriction factor that is claimed to ensure the convergence on local optima of the single non-random particle, generalizing the analytic findings to the full multi-particle system with the random weights and with the two non-stationary best values. These generalized algorithms were successfully tested on a set of benchmark functions. Some other researchers have also studied the trajectory of a single non-random particle (e.g. Kennedy et al.[3], Ozcan et al. [10], and Trelea et al. [11]).

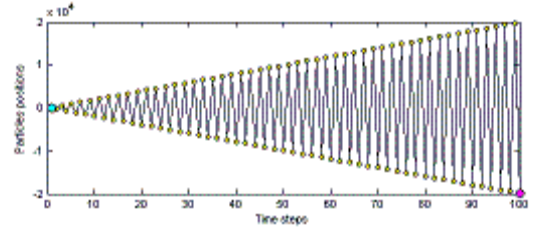

**Fig. 2**: Evolution of a single particle flying over a 1-dimensional search-space, where the two best values are fixed to zero, the particle is initially located at $x = 100$, its velocity is initialized to zero, no $v_{\max}$ is imposed, the random weights are removed, $iw = sw = 2$, and the function to be optimized is the Schaffer f6. The cyan and magenta dots are the particle's initial and final positions, respectively.

Note that although both the explosion observed in **Fig. 1** and the one observed in **Fig. 2** occur for $iw + sw = 4$, the latter is a purely deterministic explosion. While Clerc et al. [8] dealt with the mathematical reasons for this deterministic explosion, the dynamics of the explosion once the random weights $0 \leq U_{(0,1)} \leq 1$ are incorporated are not strictly considered.

If the random weights $U_{(0,1)}$ are replaced by the mean of the uniform distribution used to generate them ($\overline{U}_{(0,1)} = 0.5$), the average behaviour of the PSO according to equation **(1)** is cyclic, as shown in **Fig. 3**:

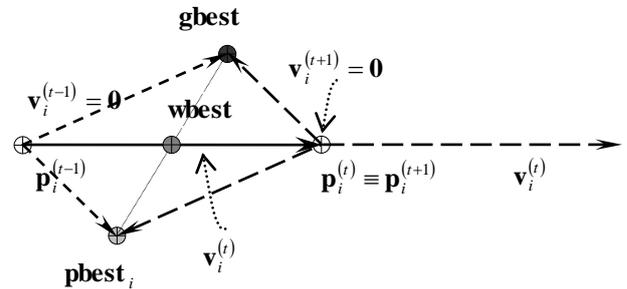

**Fig. 3**: Sketch of the trajectory of a particle $i$, which is attracted towards the points **gbest** and **pbest**$_i$, where $iw = sw = 2$, $U_{(0,1)}$ in the velocities' updating rule are replaced by $\overline{U}_{(0,1)} = 0.5$, and $\mathbf{v}_i^{(t-1)} = \mathbf{0}$. Therefore, this trajectory is in reality the part of the complete trajectory of a generic particle that is induced by the attractors at time-step $(t-1)$ (i.e. the inertia at $t-1$ is missing).

Imagine that $U_{(0,1)}$ was replaced by $\overline{U}_{(0,1)} = 0.5$ in equation **(1)**, and that the particle's velocity was initialized to 0: the particle in **Fig. 1** would move from its initial position $x = 100$ to $x = -100$ in the second time-step; it would stay in the same position in the third time-step; it would move back to $x = 100$ in the fourth time-step; and so on (note that the global optimum is located at $x = 0$ for the Schaffer f6 function). This cyclic behaviour is shown in **Fig. 4**:

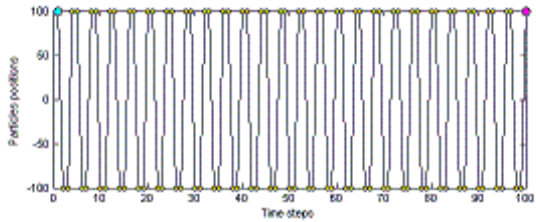

**Fig. 4**: Evolution of a single particle flying over a 1-dimensional search-space, where the two best values are fixed to zero, the particle is initially located at $x = 100$, its velocity is initialized to zero, no $v_{max}$ is imposed, $U_{(0,1)}$ is replaced by $\overline{U}_{(0,1)} = 0.5$, $iw = sw = 2$, and the function to be optimized is the Schaffer f6. The cyan and magenta dots are the particle's initial and final positions, respectively.

It is not clear why by simply incorporating the random weights instead of the constant 0.5, the particle ends up diverging rather than exhibiting a cyclic average behaviour. A simplistic heuristics argues that since the each random weight generated is as likely to be greater as it is to be less than 0.5 and there is more space to explode to than to implode to, the particle is more likely to diverge. This explosion is called "probabilistic explosion" here. In order to visualize the probabilistic explosion, the evolution of a non-random particle with $iw + sw = 0.5$ (or with $iw + sw = 1$ if the random weights $U_{(0,1)}$ are replaced by $\overline{U}_{(0,1)} = 0.5$ rather than removed) is shown in **Fig. 5**, where the deterministic explosion does not take place ($iw + sw < 4$). However, an explosion does occur as soon as the random weights are incorporated, as can be seen in **Fig. 6**.

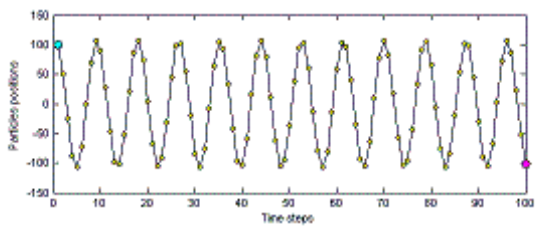

**Fig. 5**: Evolution of a single particle flying over a 1-dimensional search-space, where the two best values are fixed to zero, the particle is initially located at $x = 100$, its velocity is initialized to zero, no $v_{max}$ is imposed, $U_{(0,1)}$ is replaced by $\overline{U}_{(0,1)} = 0.5$, $iw = sw = 0.5$, and the function to be optimized is the Schaffer f6.

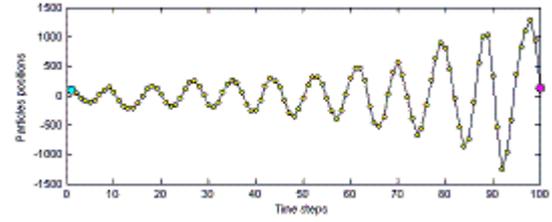

**Fig. 6**: Evolution of a single particle flying over a 1-dimensional search-space, where the two best values are fixed to zero, the particle is initially located at $x = 100$, its velocity is initialized to zero, no $v_{max}$ is imposed, $iw = sw = 0.5$, and the function to be optimized is the Schaffer f6.

As previously mentioned, clamping the components of the particles' velocities according to equation **(3)** effectively controls the explosion. Instead, Clerc et al. [8] proposed the incorporation of a constriction factor to equation **(1)**, claiming that it would ensure convergence:

$$\chi = \begin{cases} \dfrac{2 \cdot \kappa}{\left|(iw + sw) - 2 + \sqrt{(iw + sw)^2 - 4 \cdot (iw + sw)}\right|} \\ \text{if } (iw + sw) \geq 4 \\ \sqrt{\kappa} \quad \text{otherwise} \end{cases} \quad (5)$$

$$v_{ij}^{(t)} = \chi \cdot \left( v_{ij}^{(t-1)} + iw \cdot U_{(0,1)} \cdot \left( pbest_{ij}^{(t-1)} - x_{ij}^{(t-1)} \right) + sw \cdot U_{(0,1)} \cdot \left( gbest_{j}^{(t-1)} - x_{ij}^{(t-1)} \right) \right) \quad (6)$$

$$x_{ij}^{(t)} = x_{ij}^{(t-1)} + v_{ij}^{(t)} \quad (7)$$

where $\chi$ is the constriction factor and $0 < \kappa \leq 1$.

In turn, Shi et al. [12] proposed the incorporation of the inertia weight to control the explosion. Thus, the equations that rule the particles' trajectories turned into:

$$v_{ij}^{(t)} = w^{(t)} \cdot v_{ij}^{(t-1)} + iw^{(t)} \cdot U_{(0,1)} \cdot \left( pbest_{ij}^{(t-1)} - x_{ij}^{(t-1)} \right) + sw^{(t)} \cdot U_{(0,1)} \cdot \left( gbest_{j}^{(t-1)} - x_{ij}^{(t-1)} \right) \quad (8)$$

$$x_{ij}^{(t)} = x_{ij}^{(t-1)} + v_{ij}^{(t)} \quad (9)$$

where $w$ is the inertia weight.

Note that the inertia, individuality and sociality weights are not necessarily constant along the search. Given that both the original version of the algorithm and the one with the constriction factor can be viewed as particular cases of the version with the inertia weight, the latter is considered from here forth the basic PSO (B-PSO). A more general version can be thought of, where the weights are not necessarily the same for every component. However, this implies differentiating the variables from one another, leading to problem-specific rather than general-purpose optimizers.

The general flow chart of the method is shown in **Fig. 7**:

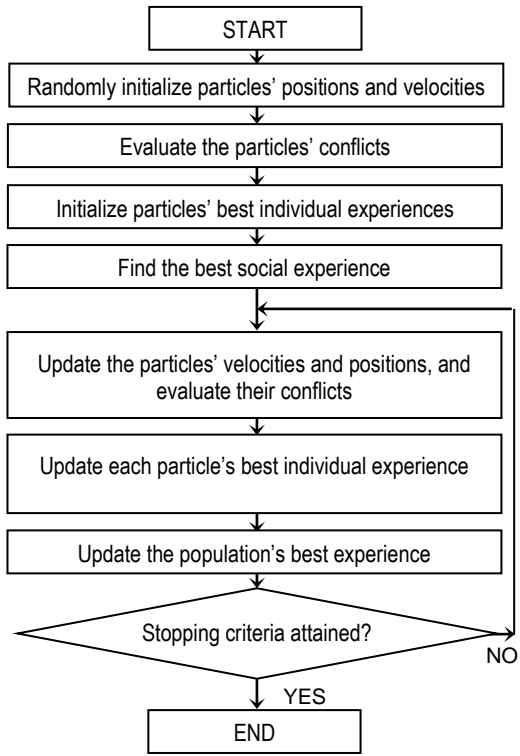

**Fig. 7**: General flow chart of the basic PSO method.

Regarding the particles which each particle interacts with, there are two main versions of the algorithm: the local PSO, and the global PSO. In the local PSO, the trajectory of a particle at a given time-step can be influenced only by its own experience and by those of a few other particles comprising its neighbourhood. Since the neighbourhoods are defined so that they overlap, the experiences can be spread over the whole population. The global version considers a single neighbourhood, so that every particle is connected to all the others. The information is spread faster in the second case. The two most common neighbourhoods are sketched in **Fig. 8**:

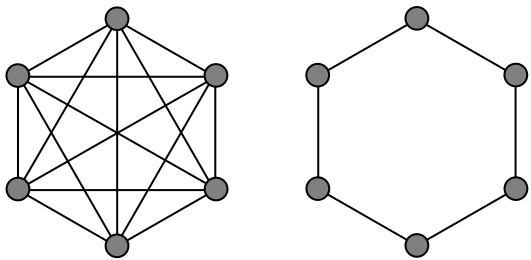

**Fig. 8**: Two typical topological neighbourhoods:

<u>Left</u>: *k*-best topology with *k* = swarm size – 1 (fully connected topology – global version)

<u>Right</u>: *k*-best topology with *k* = 2 (ring topology – local version)

This article is only concerned with the global version of the paradigm.

## PARAMETERS' TUNING

Although the constriction factor and the inertia weight are effective either in preventing the particles from exploding or at least in eventually pulling them back so that they end up clustering, it is widely agreed in the literature that the $v_{max}$ constraint should be kept. This is because it prevents subsequent evaluations of the conflict function (which can be expensive for cases such as finite element models) far from the region of interest. Several settings for the $v_{max}$ constraint were tried. Small values enhanced the fine-tuning of the search, while large values favoured exploration. However, the small values could put at risk the ability of the optimizer to escape local optima, while large values resulted in the lack of precision. Linearly time-decreasing values of $v_{max}$ were effective in enhancing the accuracy of the solutions for the original version of the PSO algorithm, but did not appear to lead to much improvement once the inertia weight was incorporated. A setting frequently found in the literature [13] is $v_{max} = 0.5 \cdot (x_{max} - x_{min})$, which is large enough not to limit the explorative behaviour yet avoids numerous evaluations of the objective function far from the region of interest: $[x_{max} - x_{min}]^n$. Thus, the enhancement of the fine-tuning of the search is left for the inertia weight and for some proposed relationships between the latter and the acceleration weight ($aw = iw + sw$).

Another important setting is the population size, which is beyond the scope of this paper. It is fair to note, however, that it is an important aspect to study, since the number of evaluations of the objective function depends on it and on the number of time-steps throughout which the search is carried out. Note that, for the same number of function evaluations, a greater population size leads to a more parallel search, while a longer search gives more time to the particles to fine-search the regions that were found to be promising. Kennedy et al. [3] suggest setting a population size between 10 and 50 particles, while Carlisle et al. [14] claim that a population size of 30 particles is a good choice. The latter is adopted here in the absence of further studies.

The beneficial effect of the (constant) inertia weight in the fine-tuning of the search can be seen in **Fig. 9** and **Fig. 10**. Several other settings for the inertia weight were implemented, keeping $iw = sw = 2$ as initially proposed by Kennedy et al. [1]. It was observed that time-decreasing inertia weights tended to favour the particles' fine-clustering—and thus, the fine-tuning of the search—even more than constant ones. It was also proposed to linearly time-swap the relative importance between the individuality and the sociality weights, keeping the acceleration weight constant, so that the particles could exhibit higher individuality at the beginning of the search and higher sociality at the end. However, the results were sometimes beneficial and sometimes harmful, turning the convenience of the strategy into problem-dependent. It seems that a strong individuality can decrease the explorative behaviour of the particles for some functions because they display smoother trajectories. Of course, a strong sociality turns the algorithm into a more local search. Hence, it seems that keeping the learning weights equal to one another—letting the random weights to dynamically alter the self-confidence of the particles—is the better choice.

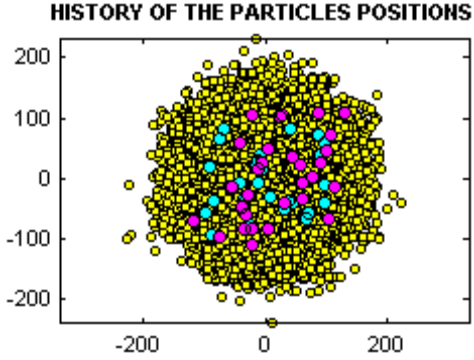

**Fig. 9**: History of the particles' positions for $w = 1$, $iw = sw = 2$, and $v_{max} = 100$ after 4000 time-steps, when optimizing the Schaffer f6 function, where the cyan and magenta dots are the initial and final particles' positions, respectively. This is equivalent to removing $w$.

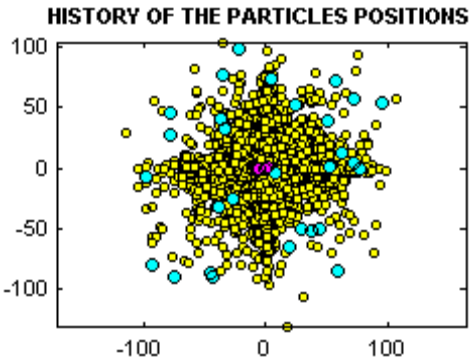

**Fig. 10**: History of the particles' positions for $w = 0.7$, $iw = sw = 2$, and $v_{max} = 100$ after 4000 time-steps, when optimizing the Schaffer f6 function, where the cyan and magenta dots are the initial and final particles' positions, respectively. Time-decreasing inertia weights result in even better fine-clustering than the one observed here.

A better strategy to enhance the particles' ability to fine-cluster is using the constricted version of the PSO, or optimizers whose inertia and acceleration weights are related like the following 4$^{th}$ degree polynomial:

$$aw^{(t)} = -4.142 \cdot \left(w^{(t)}\right)^4 + 12.398 \cdot \left(w^{(t)}\right)^3 + \\ -12.77 \cdot \left(w^{(t)}\right)^2 + 7.803 \cdot w^{(t)} + 2 = p\left(w^{(t)}\right) \quad (10)$$

This relationship was obtained by interpolating the discrete values shown in **Table 1**, which were derived from 5 geometrical analyses similar to that of **Fig. 3**, but, for each acceleration weight, a value of the inertia weight was derived so that $\mathbf{p}_i^{(t+1)} \equiv \mathbf{wbest}$ rather than $\mathbf{p}_i^{(t)} \equiv \mathbf{p}_i^{(t+1)}$ (refer to **Fig. 3**).

| *aw* | *w* |
|---|---|
| 2 | 0 |
| 3 | 1/6 |
| 4 | 1/2 |
| 5 | 9/10 |
| 6 | 4/3 |

**Table 1**: Relationship between the acceleration and the inertia weights that favours fast clustering.

Thus, 30 different optimizers were proposed, differing only in the tuning of their parameters, keeping the same $v_{max}$ and population size. They were tested on the suite of benchmark functions shown in **Table 2**, considering the average among a set of 50 runs for each experiment, so that the probabilistic nature of the method was taken into account.

The performances of the optimizers were analyzed in terms of the best solution they were able to find; of the average of the current conflicts of the particles; of its ability to, and speed of, clustering; and of its robustness.

It is important to note that, while the stronger clustering ability exhibited by some optimizers enhances the fine-tuning of the search, such ability is typically obtained in detriment of the robustness of the optimizer. Clear examples of this can be seen in **Fig. 11** and **Fig. 12**, which show the evolution of the mean[18] best and mean average[19] conflicts—represented by the Rastrigin and the Sphere functions, respectively—found by two optimizers which only differ in the parameters' setting, namely the BSt-PSO and the BSt-PSO$^{(p)}$ (refer to equations **(11)** and **(12)** in the next page for the details of the parameters' settings). The inertia and acceleration weights of the former are kept unrelated, while those of the latter are related like the 4$^{th}$-degree polynomial shown in equation **(10)**. Clearly, the BSt-PSO$^{(p)}$ exhibits a faster stagnation and an almost complete implosion of its particles, which can be inferred from the fact that the curves of its mean best and mean average conflicts virtually merge (see **Fig. 11** - right, and **Fig. 12** - right). In contrast, the BSt-PSO still exhibits a quite poor degree of clustering of its particles by the 10000$^{th}$ time-step, which can be inferred from the fact that the curves of its mean best and mean average conflicts are far from merging (see **Fig. 11** - left, and **Fig. 12** - left).

Another important observation that is in agreement with the previous conjecture about the random weights being responsible for the explosion is that the summation of the individuality and the sociality weights must be kept to less than 4 to avoid the search becoming rather random. This is observed when using the polynomial relationship, where, in spite of favouring clustering, the search appears random for inertia weights greater than 0.5 (i.e. for acceleration weights greater than 4).

---

[18] The mean is computed out of 50 runs for each experiment.

[19] The average is computed among the current conflicts of all the particles in the population.

of the algorithm), while also exhibiting high speed of clustering and a final complete implosion of the particles. It seems even less likely to find a tuning which results in fine-clustering ability on the one hand and robustness on the other. Therefore, modifications to the canonical version of the algorithm needs to be

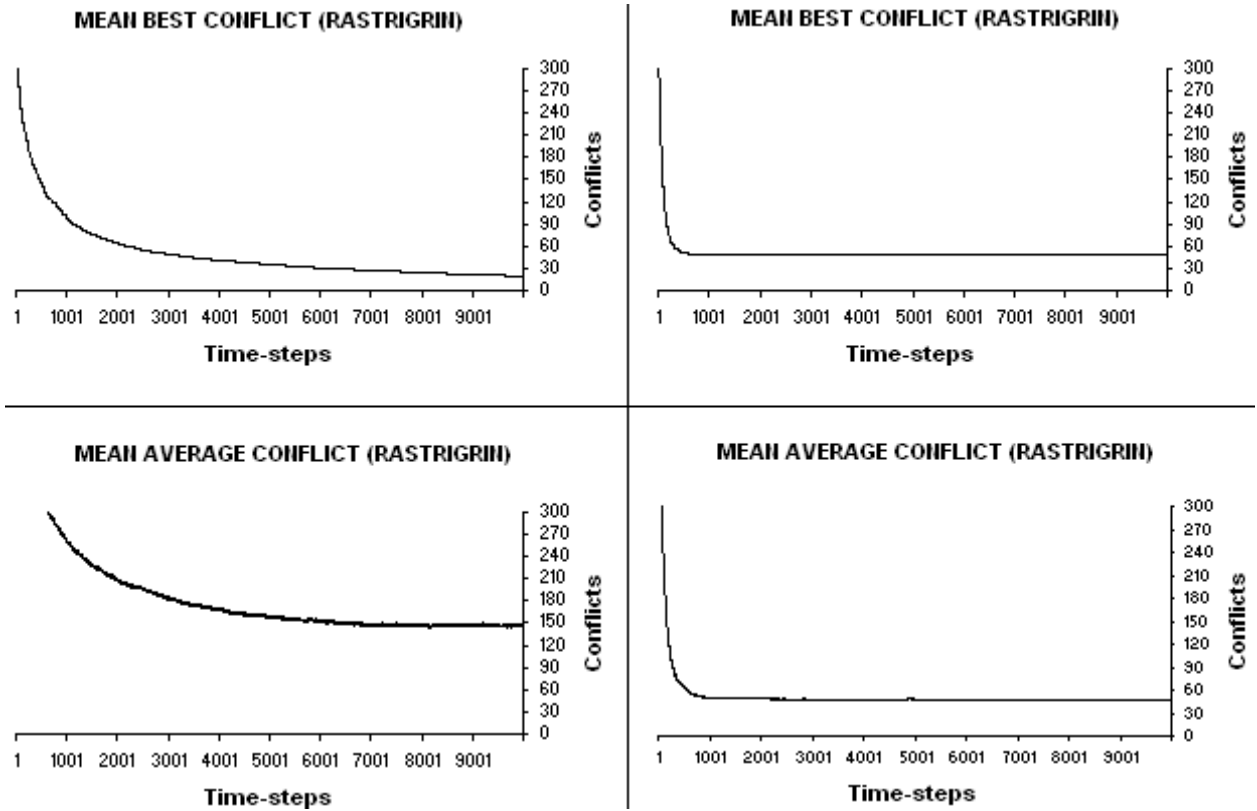

**Fig. 11**: Evolution of the mean best conflicts (above) and mean average conflicts (below) found by the BSt-PSO (left) and BSt-PSO(p) (right), where the inertia and acceleration weights of the former are unrelated and those of the latter are related like the 4th-degree polynomial, and the objective function is the 30-dimensional Rastrigrin function. The mean is computed out of 50 runs, the average among all the 30 particles, and the search is carried out along 10000 time-steps.

In summary, the acceleration weight[20] should be kept to less than 4; the optimizers that favour clustering are typically very good in optimizing functions that do not exhibit numerous local optima (such as the Sphere and Rosenbrock functions in the test suite), while those optimizers with poor fine-clustering ability are more robust in the sense of their reluctance to getting trapped in suboptimal solutions, despite not being able to fine-tune the search.

The complete experimental results of the 30 optimizers tested on the 6 functions in the test suite can be found in Innocente [15]. A brief summary is presented hereafter:

It does not seem possible to tune the parameters of the basic optimizer so that it finds the best solutions for all the 6 benchmark functions in the test suite (which were specifically included because they test different features investigated, such as subdividing the swarm in sub-swarms whose parameters are tuned so that they exhibit different characteristics.

---
[20] The constriction factor multiplied by the acceleration weight rather than the acceleration weight itself, in the case of constricted PSOs.

An example of a very robust setting, according to the experiments run on the test suite shown in **Table 2**, is as follows (**Fig. 11** - left, and **Fig. 12** - left):

BSt-PSO: $w^{(t)} = 0.7, \quad iw^{(t)} = sw^{(t)} = 2$ **(11)**

The fine-clustering ability of this optimizer is increased by using time-decreasing inertia weights, although this leads to the decrease of the explorative abilities as the search goes by.

Examples of settings that favour the fine-tuning of the search, according to the experiments run on the test suite shown in **Table 2**, are as follows:

BSt-PSO(c): $w^{(t)} = 0.7298, \quad iw^{(t)} = sw^{(t)} = 1.49609$ **(12)**

BSt-PSO(p): $w^{(t)} = 0.5, \quad iw^{(t)} = sw^{(t)} = 2$ **(13)**

The first setting is equivalent to a constricted PSO, and the second is one of those optimizers whose inertia and acceleration weights keep the 4th degree polynomial relationship shown in equation **(10)**, while also keeping the acceleration weight equal or less than 4 (**Fig. 11** - right, and **Fig. 12** - right).

These settings are also very convenient because there is no computational cost spent in complex updating rules; because they are expected to perform better when dealing with dynamic optimization problems, since their explorative ability do not decrease throughout time; and because their behaviour does not depend on the maximum permissible number of time-steps set for the search. Research on optimizers composed of different sub-swarms whose parameters are set so as to exhibit different abilities, namely robustness and the ability to fine-cluster, is currently ongoing.

computational cost can be saved. This issue is discussed along the next section.

## STOPPING CRITERIA

Traditionally, iterative methods are equipped with some stopping criteria which are met either when the solution found is good enough or when further significant improvement is unlikely. This serves the function of both saving computational cost and estimating the reliability of the solution found.

Traditional techniques, suitable for traditional methods, usually involve the difference between the best solution found up to the current time-step and that found up to the preceding one; the distance between the last two coordinates (for instance, the Euclidean norm); and a

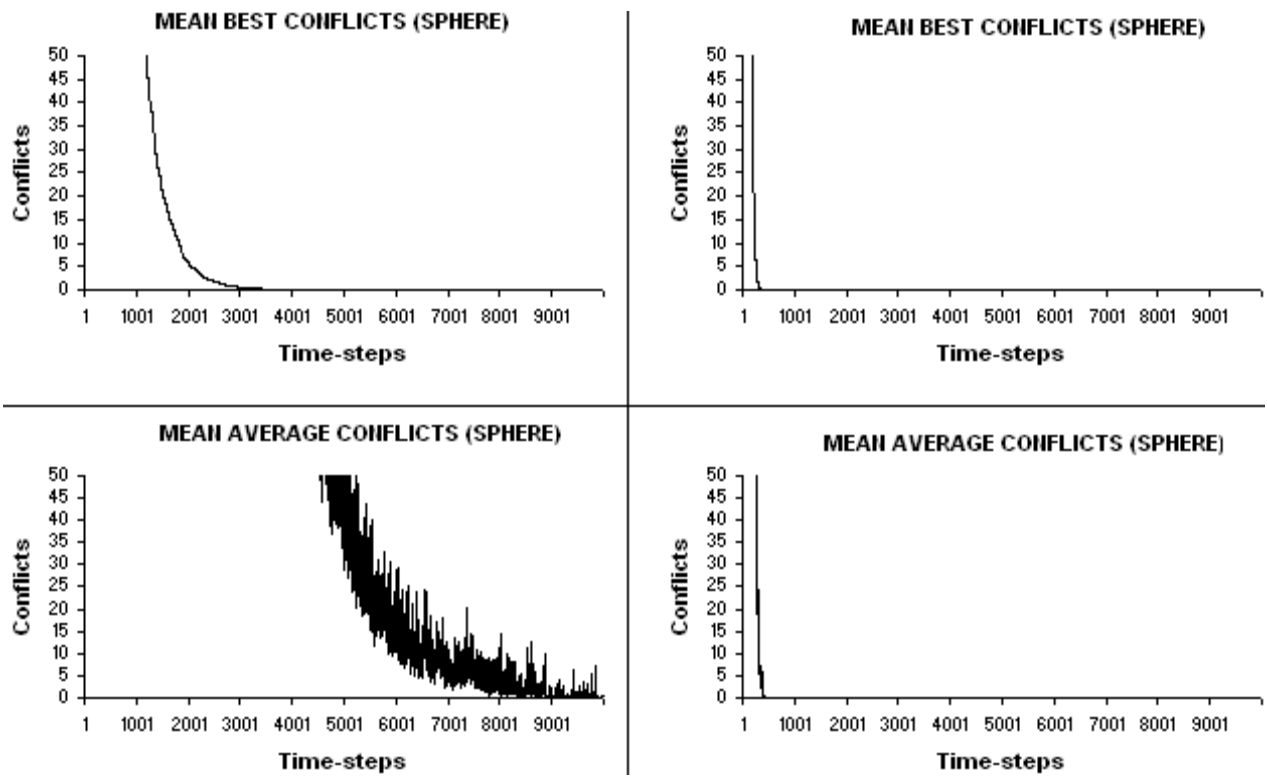

**Fig. 12**: Evolution of the mean best conflicts (above) and mean average conflicts (below) found by the BSt-PSO (left) and BSt-PSO(p) (right), where the inertia and acceleration weights of the former are unrelated and those of the latter are related like the 4th-degree polynomial, and the objective function is the 30-dimensional Sphere function. The mean is computed out of 50 runs, the average among all the 30 particles, and the search is carried out along 10000 time-steps.

A very important add-in that needs to be incorporated to the plain PSO in order to make it suitable for real-world problems is that of the stopping criteria, so that the reliability of the solution found can be estimated[21], and permissible maximum number of time-steps. However, the application of these techniques to population-based methods is not straightforward because the latter present numerous candidate solutions per time-step; the best solution found up to the current time-step might remain unchanged for quite some time before improving again; and the best solution found up to the current time-step and that found up to the preceding one might correspond to different particles, so that the distance between their locations might be an inaccurate measure of convergence.

---

[21] Although the reliability of the solution found cannot be computed for real-world problems, at least, the solution can be considered less reliable if the stopping criteria are not attained. Note, however, that it is possible to find very good solutions despite not attaining the stopping criteria.

The first, obvious aspect that has to be controlled is the evolution of the best solution found. A certain amount of improvement of the best solution found along a given number of time-steps and some arbitrary absolute errors included in the test suites of benchmark functions are the termination conditions most frequently implemented in the literature. However, these simple concepts are not sufficient for a general-purpose optimizer which aims to be applicable to real-world problems: the absolute errors are problem-specific, and their permissible values are difficult to be set when the solution to the problem is unknown; and the threshold of the rate of improvement below which the improvement is considered negligible represents different degrees of importance for different conflict functions.

Traditional measures of error used to develop stopping criteria typically complement the errors computed with regards to the objective function with those computed with regards to the coordinates' values. The reason for this is that the function to be optimized may exhibit small differences in its evaluation at coordinates which are far from one another, or, in contrast, it may exhibit great differences in its evaluation at coordinates which are very close to each other. Examples of the first case are functions that present extensive flat areas, while a clear example of the second case is the Schaffer f6 function (see **Fig. 13**). Thus, it is reasonable to design stopping criteria involving measures of error with regards to the conflict values on the one hand, and measures of error with regards to the particles' positions on the other.

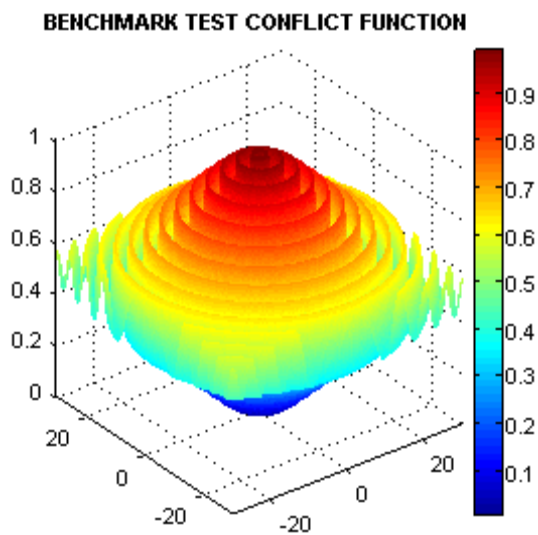

**Fig. 13**: Surface plot of the Schaffer f6 function for 2-dimensional search-spaces in the region $[-30,30]^2$.

In addition, the measures of error—either regarding the conflict values or the particles' positions—should take into account both the evolution of the conflicts and coordinates throughout the time-steps, and the degree of clustering of the particles within a single time-step. This is because the best solution found might remain unchanged for quite some time without that necessarily implying that further significant improvement is unlikely. While the particles still exhibit a poor degree of clustering, further improvement is possible. Once the particles have achieved a high degree of clustering, the chances of further improvement decrease dramatically. However, improvement may still be achieved with the whole swarm behaving very much like a single particle (refer to **Fig. 11** - right, and **Fig. 12**- right), although this implies a drastic decrease in the algorithm's robustness.

Last but not least, there might be some problems such as the Rosenbrock function in our experiments, where the particles of some optimizers with the ability to fine-cluster display a small explosion after an initial high degree of clustering (refer to **Fig. 14**). Although the reason for this is not clear, this might result in not attaining the termination conditions despite finding very good solutions.

Two sets of termination conditions were developed, where attaining any of them leads to the termination of the search. The first set considers the degree of clustering of the particles; a low threshold for the rate of improvement of the best solution, below which the improvement is believed to be negligible; and both a minimum and a maximum number of time-steps permitted for the search to go through. In contrast, the second set of termination conditions does not consider the degree of clustering but sets a very demanding threshold for the maximum permissible rate of improvement of the best solution: the search is terminated if there is no improvement for a 35% of the maximum number of time-steps permitted for the search to go through. In addition, the second set of termination conditions also includes a minimum and a maximum number of time-steps.

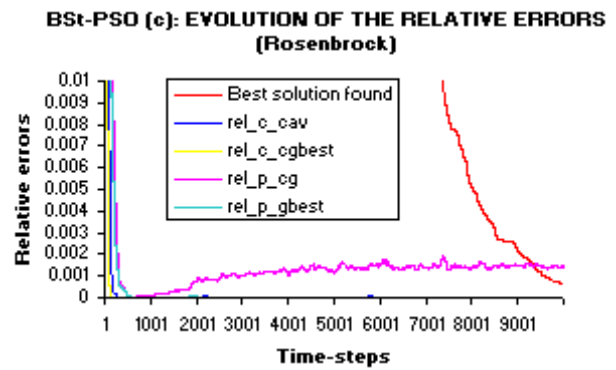

**Fig. 14**: Evolution of the relative errors designed between consecutive time-steps for the BSt-PSO[(c)] optimizing the 30-dimensional Rosenbrock function.

The measures of error considered for the design of the stopping criteria to be incorporated into general-purpose optimizers must be independent from the conflict function, from the number of design variables, from the size of the search-space, and from the number of particles in the swarm. Therefore, relative errors are preferred over absolute ones. However, it is not straightforward to decide on a convenient value to relate the absolute errors to. The first values that come to mind are the best solution found so far and its location, so that the permissible relative errors can be set as a percentage of the true solution and of its coordinates[22]. However, this strategy does not work properly when the optimum is near or equal to zero. In addition, it seems reasonable to argue that the error should be limited to a percentage of the maximum error possible rather than to a percentage of the true solution. For instance, consider the Schaffer f6 function (refer to **Fig. 13**) modified by the addition of a very big constant, say 10000. A candidate solution equal to 10000.5 results in a relative error equal to $5 \times 10^{-5}$, which appears to be acceptable at first glance. However, since this function oscillates within the interval $[10000, 10001]$, even a random solution is likely to exhibit such an error. In fact, the maximum relative error possible is less than $1 \times 10^{-4}$. Therefore, it is proposed here to relate the absolute errors regarding the conflict values to the difference between the best and the worst solutions that the algorithm is able to find along the whole search. In order to make this possible, a specialized sub-swarm composed of only five particles is added to the population, which is in quest for the worst rather than for the best conflict. Because the worst conflict is only used for the computation of the relative errors, high precision is not essential. Likewise, the errors regarding the particles' positions are related to the size of the feasible search-space: $x_{max} - x_{min}$.

Notice that this strategy is only possible for problems where the search-space is constrained to a finite region. Hence the "preserving feasibility" technique is brought to this section to handle hyper-cube-like boundary constraints. This is a very robust constraint-handling technique that consists of successively initializing the particles randomly until the whole population is spread over the feasible space, and thereafter simply banning from memory the infeasible solutions. This technique is briefly discussed in the next section.

The curves of the evolution of measures of error which involve numerous individuals whose interrelations are stochastically weighted present rough shapes with wide and uneven oscillations. This makes their use in the design of stopping criteria quite difficult. In addition to that, while the quantitative analysis of the proposed measures were performed considering the average of the results obtained from 50 runs for each experiment, the average curves are always smoother than those corresponding to a single run. Therefore, a qualitative analysis involving a single run was also performed, obtaining very rough curves with wide and uneven oscillations. This problem was solved by involving the last 100 time-steps in the computation of the current measures of error, as it can be seen in the proposed equations hereafter:

**First set of termination conditions**

The search is terminated if the following 8 conditions are met:

1) $t \geq 0.1 \cdot t_{max}$

2) $rel\_c\_me^{(t)} = \dfrac{\sum_{i=t-99}^{t}\left(\overline{c}^{(i)} - cgbest^{(i)}\right)}{100 \cdot \left(cgworst^{(t)} - cgbest^{(t)}\right)} \leq 10^{-12}$

3) $rel\_p\_mse^{(t)} = \dfrac{\sum_{i=t-99}^{t}\sqrt{\sum_{j=1}^{m}\sum_{k=1}^{n}\left(x_{jk}^{(i)} - gbest_{k}^{(i)}\right)^2}}{100 \cdot (x_{max} - x_{min}) \cdot \sqrt{m \cdot n}} \leq 10^{-9}$

4) $rel\_p\_cg\text{-}gbest^{(t)} =$
$= \dfrac{\sum_{i=t-99}^{t}\sqrt{\sum_{j=1}^{n}\left(cg_{j}^{(i)} - gbest_{j}^{(i)}\right)^2}}{100 \cdot (x_{max} - x_{min}) \cdot \sqrt{n}} \leq 10^{-9}$

5) $rel\_c\_cav^{(t)} = \dfrac{abs\left(\overline{c}^{(t)} - \overline{c}^{(t-100)}\right)}{100 \cdot \left(cgworst^{(t)} - cgbest^{(t)}\right)} \leq 10^{-12}$

6) $rel\_c\_cgbest^{(t)} = \dfrac{cgbest^{(t-100)} - cgbest^{(t)}}{100 \cdot \left(cgworst^{(t)} - cgbest^{(t)}\right)} \leq 10^{-15}$

7) $rel\_p\_cg^{(t)} = \dfrac{\sum_{i=t-99}^{t}\sqrt{\sum_{j=1}^{n}\left(cg_{j}^{(t)} - cg_{j}^{(t-1)}\right)^2}}{100 \cdot \sqrt{n} \cdot (x_{i\,max} - x_{i\,min})} \leq 10^{-9}$

8) $rel\_p\_gbest^{(t)} =$
$= \dfrac{\sum_{i=t-99}^{t}\sqrt{\sum_{j=1}^{n}\left(gbest_{j}^{(t)} - gbest_{j}^{(t-1)}\right)^2}}{100 \cdot \sqrt{n} \cdot (x_{i\,max} - x_{i\,min})} \leq 10^{-12}$

**Second set of termination conditions**

The search is terminated if the following two conditions are met:

1) $t > 0.35 \cdot t_{max}$

2) $cgbest^{(t-0.35 \cdot t_{max})} - cgbest^{(t)} = 0$

Where:

- $c_i^{(t)}$ : conflict of particle *i* at time-step *t*

- $\overline{c}^{(t)}$ : average among the conflicts of the particles in the swarm at time-step *t*

---

[22] This assumes that the best solution found so far equals the exact solution, which is just an approximation.

- $m$ : number of particles in the swarm
- $n$ : number of design variables
- $t_{max}$ : maximum number of time-steps permitted
- $cgbest^{(t)}$ : best solution found so far
- $cgworst^{(t)}$ : worst solution found so far
- $gbest_k^{(i)}$ : $k^{th}$ coordinate of the best solution found up to time-step $i$
- $cg_j^{(i)}$ : $j^{th}$ coordinate of the centre of gravity of the swarm at time-step $i$
- $(x_{max} - x_{min})$ : feasible range of the search-space

Notice that terminating the search due to the attainment of one or the other set of conditions has completely different implications. Attaining the first set of conditions implies that the particles have achieved a high degree of clustering, and that the rate of improvement of the solution has reached a lower permissible threshold. Fulfilling the second set of terminations conditions implies that, although the particles have not yet achieved the required degree of clustering, further improvement of the best solution found appears unlikely.

It is important to remark that attaining either one or the other set of conditions does not give direct information with regards to the goodness of the solution found. For instance, the second set of conditions might be met by an optimizer which is not able to improve the solutions because its particles do not cluster at all. However, this is a problem of designing the algorithm itself rather than a problem of the design of the stopping criteria.

The stopping criteria was then incorporated into the optimizer, and the BSt-PSO[c] and the BSt-PSO[p] were tested on the suite of benchmark functions shown in **Table 2**. Some of the most relevant results are gathered in **Table 3**. The maximum number of time-steps was set to 30000, and the second set of error conditions was slightly modified, reducing the constant 0.35 to 0.25. It can be observed that both optimizers achieve a high degree of clustering when optimizing the Sphere, Rastrigin and Griewank functions. In fact, the search is terminated at the 3000[th] time-step, which is the earliest possible. However, they both appear to get trapped in a local optimum when optimizing the Rastrigin function, and the BSt-PSO[c] also when optimizing the Griewank function. They do not have any trouble either in performing the implosion or in finding the exact solution when optimizing the 2-dimensional Schaffer f6 function, although they take more than 3000 time-steps to do so. However, the 30-dimensional Rosenbrock and Schaffer f6 functions appear considerably harder to be optimized, and they either attain the second set of termination conditions or none at all. While the only case where none of the sets of termination conditions is attained is that of the BSt-PSO[c] optimizing the Rosenbrock function, a very good solution is found. This is because the best solution found does not stop improving and the particles do not attained the required degree of clustering!

A rather academic experiment was carried out in order to show the smooth shape of the proposed measures of error, and the two clusters of particles formed, one around the best, and another around the worst solution found within the feasible space. Thus, the 2-dimensional Sphere function was optimized by the BSt-PSO[p], setting the maximum number of time-steps permitted for the search to go through equal to 10000.

Of course, this is a very simple problem, and the optimizer found no difficulty in obtaining the global optimum very quickly. The evolution of the proposed relative errors is shown in **Fig. 15** (regarding the conflict values) and **Fig. 16** (regarding the particles' positions). The history of the particles' positions in the search-space is shown in **Fig. 17**.

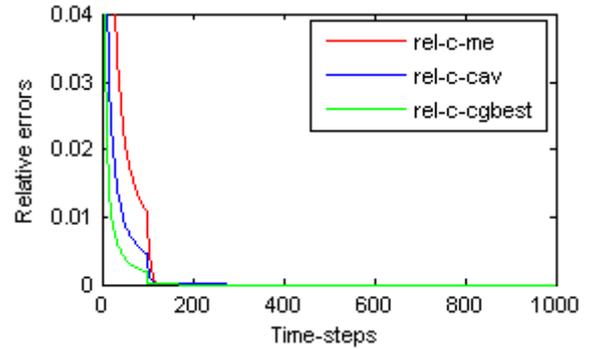

**Fig. 15**: Evolution of the relative errors regarding the conflict values for the BSt-PSO[p] optimizing the 2-dimensional Sphere function, where the feasible search-space is given by the hyper-cube $[-100,100]^2$, and the particles are initialized within the hyper-cube $[-75,-25]^2$.

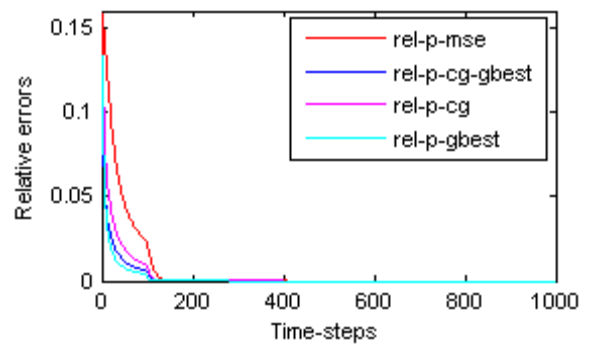

**Fig. 16**: Evolution of the relative errors regarding the particles' positions for the BSt-PSO[p] optimizing the 2-dimensional Sphere function, where the feasible search-space is given by the hyper-cube $[-100,100]^2$, and the particles are initialized within the hyper-cube $[-75,-25]^2$.

The feasible search-space was set to $[-100,100]^2$, while the particles were initialized within the region $[-75,-25]^2$ in order to facilitate the visualization of the two sub-swarms splitting: one seeking the best and the other seeking the worst solution possible within the feasible search-space. Note that the particles are allowed to fly over infeasible search-space (see **Fig. 17**).

Notice that in the particular case of the BSt-PSO[(c)] optimizing the Rosenbrock function—where the small divergence takes place—, the best solution found never stops improving, and the degree of clustering does not meet the stopping criteria. Hence the search is never stopped despite the fact that the algorithm finds a very good solution. It is important to keep in mind that the stopping criteria is more concerned with the likelihood

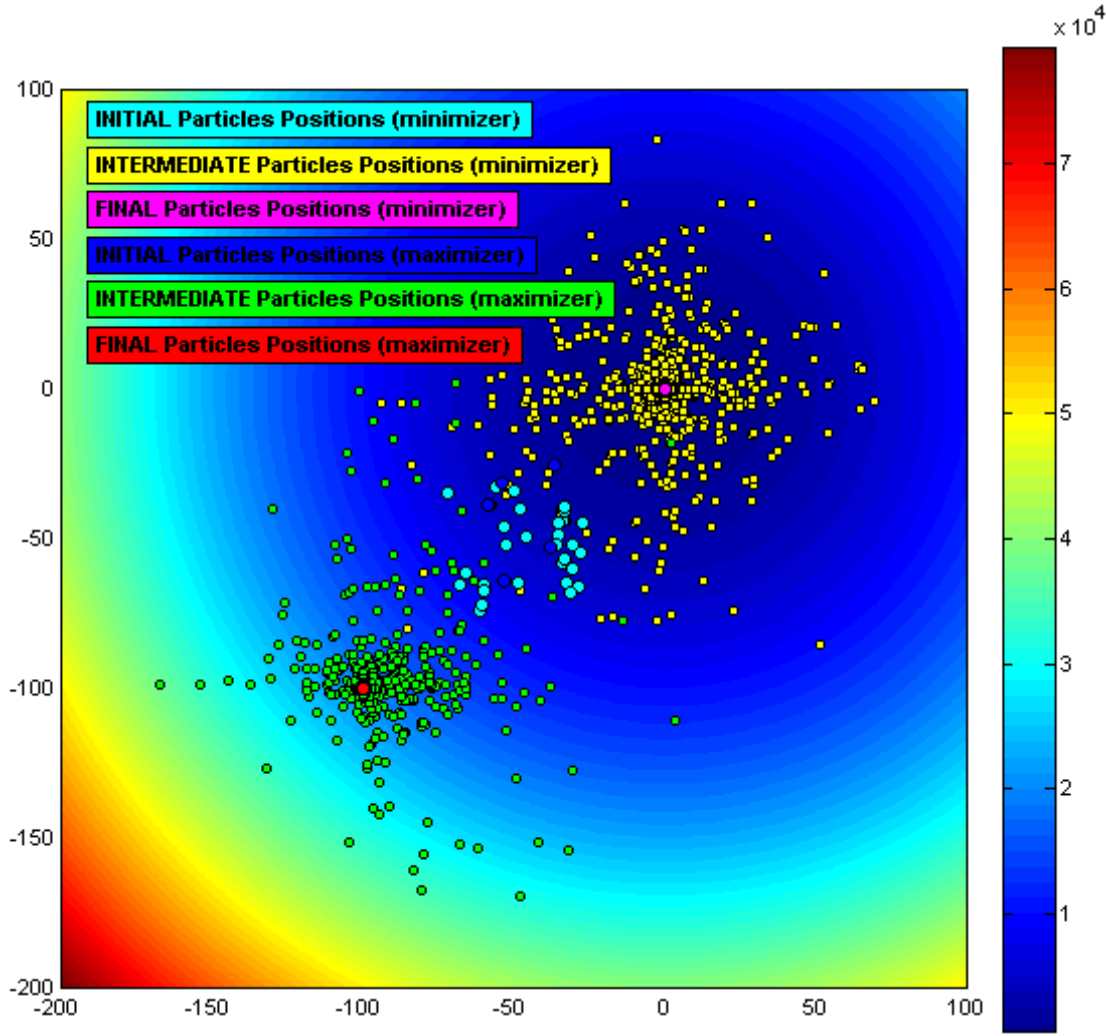

**Fig. 17**: Evolution of the particles' positions for the BSt-PSO[(p)] optimizing the 2-dimensional Sphere function, where the feasible search-space is given by the hyper-cube $[-100,100]^2$, the particles are initialized within the hyper-cube $[-75,-25]^2$, 30 particles are in quest for the best solution possible (minimizer), and 5 particles are in quest for the worst solution possible (maximizer), which is necessary to compute the proposed the relative errors. Notice that the particles are allowed to fly over infeasible search-space.

It is interesting to observe that the solutions found when the termination conditions are attained (refer to **Table 3**) also satisfy the acceptable exact absolute errors stated in **Table 2**, whose first 5 functions and their permissible values were taken from Carlisle et al. [14][23].

---

[23] Notice that the absolute error conditions stated in **Table 2**, which can be set because the global optimum is well known for benchmark functions, were not used at all for the development of the proposed stopping criteria.

of finding significantly better solutions rather than with the reliability of the solution found. It is the design of the optimizer itself which should be concerned with the reliability of the solution it is able to find before stagnating. It has been preferred here to eventually "waste" computational resources in maintaining the search running in spite of finding a good solution rather than to eventually terminate the search while the solution is still unacceptable. Therefore, a first attempt for the development of a general-purpose optimizer is proposed, which intends to combine the robustness that results from some settings with the fine-clustering ability that results from some others.

Thus, the settings for the proposed general-purpose PSO (GP-PSO) are as follows:

A swarm of 30 particles seek the global minimum, where the parameters of the velocity updating equation of 10 particles are as follows:

✗  $w^{(t)} = 0.7, \ iw^{(t)} = sw^{(t)} = 2$

The velocity of other 10 particles is updated according to the following settings:

✗  $w^{(t)} = 0.5, \ iw^{(t)} = sw^{(t)} = 2$

The trajectory of the last 10 particles of the minimizer is ruled by the following parameters:

✗  $w^{(t)} = 0.7298, \ iw^{(t)} = sw^{(t)} = 1.49609$

Finally, the maximizer is composed of other 5 particles which seek the worst solution possible within the feasible search-space. The parameters' setting for the maximizer is as follows:

✗  $w^{(t)} = 0.7, \ iw^{(t)} = sw^{(t)} = 2$

It is fair to note that all the particles are fully connected, and the evaluation of the conflict of any particle can become either the best or the worst solution found. For instance, a particle of the minimizer can accidentally find a solution that is worse than any of the solutions found so far by the particles of the maximizer. Then, the particles of the maximizer can be attracted towards a location found by a particle of the minimizer (while the particle that found such location is not attracted towards it!), and vice versa. A good example for this is that of the Schaffer f6 function, where the best and worst solutions are located near one another (see **Fig. 13**).

The most relevant results obtained from testing this GP-PSO are gathered in **Table 3**. In addition, the evolution of the errors regarding the conflict values can be seen in **Fig. 19**, and the one of the errors regarding the particles' positions in **Fig. 20**. As expected, the different features that results from the different settings were successfully combined, and this optimizer encompasses all the beneficial features of the three settings it is composed of (compare the results of the experiments corresponding to the three optimizers in **Table 3**). However, the problem of the Rosenbrock function is still there[24]. It is important to remark that the stopping criteria proposed here comprises just a first attempt, and further research and work are necessary for improvement.

The next step towards a general-purpose optimizer is the incorporation of a robust constraint-handling technique.

## CONSTRAINT-HANDLING TECHNIQUES

Although different techniques have been proposed in the literature to deal with the constraints, the appropriate choice appears to be problem-dependent. An extensive discussion on the different existing techniques and their variations is beyond the scope of this paper. It is only intended here to make a brief review of the concepts behind the most popular ones, including the "preserving feasibility" technique implemented in the GP-PSO, whose performance was tested on a suite of benchmark functions in previous sections. Further tests on more complex constrained optimization problem are required.

Broadly speaking, three main groups of constraint-handling techniques appropriate for particle swarm optimizers can be differentiated[25]:

1. "Cut off" technique
2. "Preserving feasibility" technique
3. "Penalization" technique

It is fair to note that these techniques are suitable for inequality constraints. Although some adaptations can be performed to handle equality ones, such adaptations are not discussed in this paper.

### "Cut off" technique

The simplest version of this technique is straightforward for hyper-cube-like boundary constraints. That is, when the design variables present a continuous range of feasible values. Its formulation is similar to the $v_{\max}$ constraint to the components of the particles' velocities:

**if** $\quad x_{ij}^{(t)} > x_{\max} \ \Rightarrow \ x_{ij}^{(t)} = x_{\max}$

**elseif** $\quad x_{ij}^{(t)} < x_{\min} \ \Rightarrow \ x_{ij}^{(t)} = x_{\min}$ (14)

Where $[x_{\min}, x_{\max}]$ is the range of feasible values of the design variables.

Note that this results in placing the particle somewhere on the boundary, and in the direction of the velocity being altered. An alternative is to implement a subroutine to find the intersection between the original velocity vector (i.e. the increment of displacement) and the boundary, and locate the particle there. In our experiments, there did not seem to be much difference between these two alternatives, although only simple problems and single runs were carried out, so that final conclusions cannot be made. The "cut off" technique appears to be efficient when the solution is located somewhere on the boundary.

---

[24] The problem is that the improvement never stops, and the complete implosion of the particles seems to never take place. Hence the termination conditions are not attained despite the very good solutions found.

[25] The denomination and particular features of these techniques are not uniform. Different denominations may refer to similar techniques and equal denominations may refer to different techniques (e.g. the denomination "preserving feasibility" refers to different techniques for different researchers).

Some other modifications include reflection rather than "cut off", which is expected to improve the performance of the algorithm when the solution is near but not on the boundary. Some interesting forms of reflections are proposed by Foryś et al. [16].

**"Preserving feasibility" technique**

The implementation of this technique is straightforward, requiring very few variations to the plain unconstrained optimizer. In fact, the algorithm is kept the same as if the problem was unconstrained. Thus, the particles are allowed to fly over infeasible space, as opposed to the "cut off" technique. The only modification with respect to the unconstrained algorithm is the incorporation of a condition on the subroutine of the update of each particle's best previous experience: if a constraint is violated, the candidate solution cannot become a best experience, regardless of the value of the conflict function associated to that position. It is evident that this strategy requires that all the particles are initialized within the feasible space. This is typically performed by brute force, by repeatedly and randomly initializing each particle until the whole population is feasible.

Although this is a robust strategy, it may be inefficient and the initialization may fail when the feasible search-space is small in size; when it is composed of disjointed sub-spaces; and/or when the size of the population is too big. This technique was proposed by Hu et al. [17, 18].

The influence that the "cut-off" and the "preserving feasibility" strategies have on the behaviour of the swarm is illustrated in **Fig. 18**, where the Sphere function is optimized, and the feasible search-space is delimited by the region $[50, 250]^2$.

Notice that both searches were performed along 4000 time-steps. It seems that the "cut off" technique is faster for the solutions located on the boundary, but it exhibits a noticeably poorer exploration of the search-space.

**"Penalization" technique**

This method is a standard procedure to deal with constraints in EAs. Similar to the "preserving feasibility technique", the idea is to turn the constrained problem into an unconstrained one, so that the inherently unconstrained optimization method can deal with it.

Thus, the particles searching the infeasible space are evaluated, but their conflicts are increased if the solution is infeasible.

$$fp(\mathbf{x}) = f(\mathbf{x}) + Q(\mathbf{x}) \quad (15)$$

Where:
- $fp(\mathbf{x})$: penalized fitness of particle $\mathbf{x}$.
- $f(\mathbf{x})$: fitness of particle $\mathbf{x}$.
- $Q(\mathbf{x})$: penalty for infeasible particle $\mathbf{x}$.

Often, penalties are not fixed but linked to the amount of infeasibility of the individual. They might simply be functions of the number of constraints violated, but functions of the distance from feasibility are usually preferred. For instance, for optimization problems of the form:

$$\text{Minimize } f(\mathbf{x})$$
$$\text{with } \mathbf{x} \in \mathcal{R}^n \quad (16)$$

Where:
- $g_j(\mathbf{x}) \leq 0 \quad ; \quad j = 1, \ldots, q$
- $g_j(\mathbf{x}) = 0 \quad ; \quad j = q+1, \ldots, m$

The degrees of infeasibility might be taken into account by constraints violation measures:

$$f_j(\mathbf{x}) = \begin{cases} \max\{0, g_j(\mathbf{x})\} & ; \quad 1 \leq j \leq q \\ g_j(\mathbf{x}) & ; \quad q < j \leq m \end{cases} \quad (17)$$

Therefore, the corrected conflict value is as follows:

$$fp(\mathbf{x}) = f(\mathbf{x}) + \lambda(t) \cdot \sum_{j=1}^{m} (f_j(\mathbf{x}))^2 \quad (18)$$

Where $\lambda(t)$ is updated every generation according to:

$\lambda(t+1) = \dfrac{1}{\beta_1} \cdot \lambda(t)$, if the best particle in the last $k$ generations was always feasible.

$\lambda(t+1) = \beta_2 \cdot \lambda(t)$, if the best particle in the last $k$ generations was never feasible. (19)

$\lambda(t+1) = \lambda(t)$, otherwise.

Where $\beta_1, \beta_2 > 1 \wedge \beta_1 \neq \beta_2$. Note that if
$g_j(\mathbf{x}) \leq 0 \ \forall j = 1, \ldots, q \ \wedge \ h_j(\mathbf{x}) = 0 \ \forall j = q+1, \ldots, m$
$\Rightarrow \quad f_j(\mathbf{x}) = 0 \ \forall j \quad \Rightarrow \quad fp(\mathbf{x}) = f(\mathbf{x})$.

The penalization method is a very popular technique. However, the tuning of the parameters of the method is not an easy task. A high penalization might lead to the situation where the particles cannot search the infeasible regions, thus converging to a sub-optimal but feasible solution. A low penalization might lead to a system where the particles are violating constraints but present themselves as fitter than feasible individuals. The proper definition of the penalty functions is not trivial, and it plays a crucial role in the performance of the algorithm. A penalization method similar to the one discussed here is proposed by Venter [19], while Konstantinos et al. [20] propose a more sophisticated one.

Several variations can be made to these constraint-handling techniques, such as resetting the velocity of the particle to zero if the latter is flying over infeasible space. This removes the effect of the inertia, and the particle is pulled back to the feasible region faster.

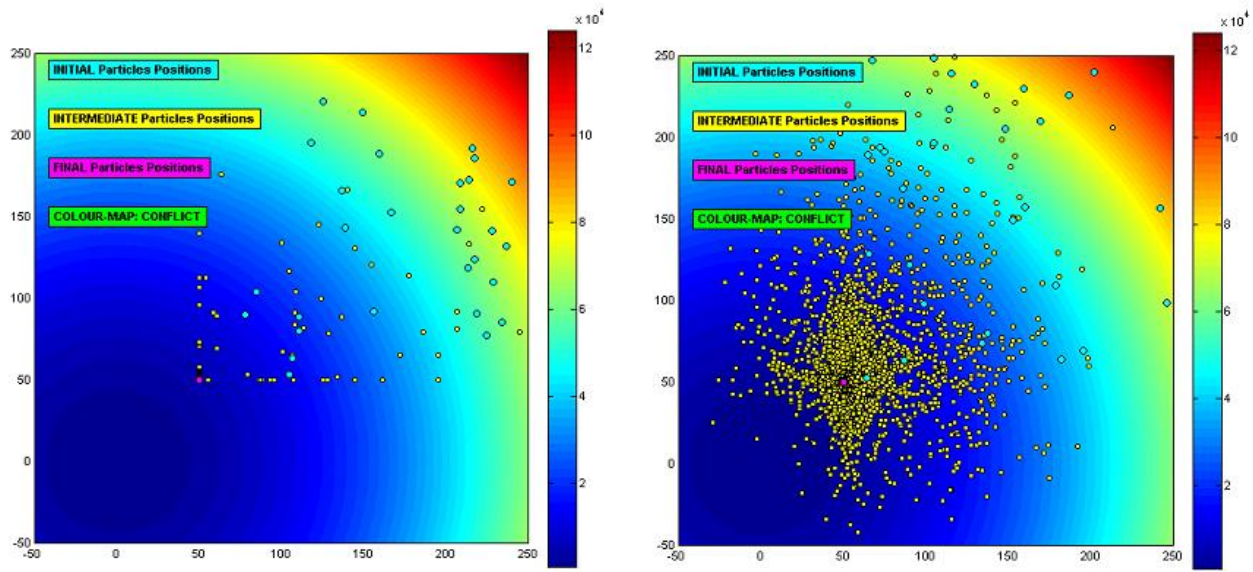

**Fig. 18**: A PSO algorithm optimizing the 2-dimensional Sphere function, where the feasible part of the search-space is delimited by $[50,250]^2$. The constraint-handling techniques are the "cut off" (left) and the "preserving feasibility" technique (right). The search is carried out along 4000 time-steps in both cases.

Basic experiments on these three main techniques to handle constraints were performed on very simple constrained optimization problems (mainly on problems with hyper-cube-like boundary constraints). No final conclusion could be derived in the sense of a convenient general-purpose constraint-handling method. However, this research is in its early stages, and further work on this matter is currently ongoing.

The "preserving feasibility" technique was chosen for this GP-PSO in spite of its weaknesses because it is, in principle, suitable for any kind of inequality constraint, it requires few modifications to the basic unconstrained algorithm, and no parameter needs to be tuned.

## CONCLUDING REMARKS

The PSO is a robust, general-purpose method whose original version does not include convergence criteria or constraint-handling techniques. In addition, the original parameters' tuning resulted in optimizers incapable of fine-tuning the search. Therefore, some tunings have been proposed and tested, and a general-purpose optimizer (GP-PSO) was developed combining different desirable features of some homogeneous swarms tested before. Some measures of error that are appropriate for particle swarm optimizers were designed, incorporated to the algorithm, and tested with promising results.

Finally, a few constraint-handling methods were briefly discussed, although the "preserving feasibility" technique had been already implemented when testing the stopping criteria. Thus, a first general-purpose particle swarm optimizer (GP-PSO) was developed, which is expected to handle real-world problems.

## FUTURE WORK

Although the proposed GP-PSO is already an optimizer well able to deal with real-world problems, it is still in its early stages. With regards to the unconstrained algorithm, settings for the velocity constraint other than $v_{max} = 0.5 \cdot (x_{max} - x_{min})$ should be tested. In addition, further study of the influence of the random weights needs to be carried out, so that other improvements to the basic algorithm can be considered. For instance, generating the random weights from a Gaussian rather than from a uniform distribution might result in faster convergence; or using the same random weight for all the coordinates of a particle in the velocity updating rule might also result in faster convergence. While it has been concluded that it is better to set the individuality and sociality weights equal to one another thus leaving the random weights alternate the relative importance of the learning weights dynamically and stochastically, perhaps the relative importance between the inertia and the acceleration weights should also be altered dynamically and stochastically by multiplying the inertia weight by a random weight between 0 and 2.

Another aspect related to the metaphor that inspired the method is that of learning in three levels: individually, socially, and from culture. In the global PSO, it could be considered that a particle learns from its own experience and from experiences of particles that it might have never even met by means of culture. Therefore, the learning by observation of the neighbours' behaviour is missing. In the local PSO, the particle learns from its neighbours, but the learning from the culture is missing.

Trying the three-level learning might be worth trying. In addition, the enhancement of the individual learning by means of a local search will be tried in the near future.

Other techniques such as updating the best experiences every time a particle's position is updated rather than doing it in parallel, and local versions of the algorithm are also interesting aspects to investigate further. The local version has been tried showing little difference with respect to the global version. However, it is reasonable to expect that the local version would work at its best when optimizing a function that displays numerous local optima located far from one another, as opposed to the benchmark functions in the test suite.

The influence of the population size is very important because it has a direct impact on the number of evaluations of the conflict function, which can be quite expensive in cases such as that of finite element modelling. Thus, the influence of the population size and a detailed study of the computational costs of the optimizer should also be carried out in the future.

|  | Mathematical expression | Parameters |
|---|---|---|
| **Sphere** | $f(\mathbf{x}) = \sum_{i=1}^{n} x_i^2$ | - Search-space: $[-100,100]^{30}$<br>- Acceptable error: < 0.01 |
| **Rosenbrock** | $f(\mathbf{x}) = \sum_{i=1}^{n-1} 100 \cdot (x_{i+1} - x_i^2)^2 + (x_i - 1)^2$ | - Search-space: $[-30,30]^{30}$<br>- Acceptable error: < 100 |
| **Rastrigrin** | $f(\mathbf{x}) = \sum_{i=1}^{n} [x_i^2 - 10 \cdot \cos(2 \cdot \pi \cdot x_i) + 10]$ | - Search-space: $[-5.12, 5.12]^{30}$<br>- Acceptable error: < 100 |
| **Griewank** | $f(\mathbf{x}) = \frac{1}{4000} \cdot \sum_{i=1}^{n} x_i^2 - \prod_{i=1}^{n} \cos\left(\frac{x_i}{\sqrt{i}}\right) + 1$ | - Search-space: $[-600,600]^{30}$<br>- Acceptable error: < 0.1 |
| **Schaffer f6 2D** | $f(\mathbf{x}) = \frac{\left(\sin\sqrt{x_1^2 + x_2^2}\right)^2 - 0.5}{[1 + 0.001 \cdot (x_1^2 + x_2^2)]^2} + 0.5$ | - Search-space: $[-100,100]^{2}$<br>- Acceptable error: < 0.00001 |
| **Schaffer f6** | $f(\mathbf{x}) = \frac{\left[\sin\left(\sqrt{\sum_{i=1}^{n} x_i^2}\right)\right]^2 - 0.5}{\left(1 + 0.001 \cdot \sum_{i=1}^{n} x_i^2\right)^2} + 0.5$ | - Search-space: $[-100,100]^{30}$<br>- Acceptable error: < 0.1 |

**Table 2**: Benchmark functions in the test suite. The acceptable absolute errors are applicable for the experiments performed prior to the development of the stopping criteria.

| | BSt-PSO[c] | | | BSt-PSO[p] | | | GP-PSO | | |
|---|---|---|---|---|---|---|---|---|---|
| **FUNCTION** | Solution | Time-steps to meet stopping criteria | Set of termination conditions attained | Solution | Time-steps to meet stopping criteria | Set of termination conditions attained | Solution | Time-steps to meet stopping criteria | Set of termination conditions attained |
| Sphere | 1.17E-45 | 3000 | 1 | 1.19E-37 | 3000 | 1 | 6.77E-49 | 4155 | 1 |
| Rosenbrock | 6.48E-10 | - | - | 3.70E+01 | 17829 | 2 | 7.89E-10 | - | - |
| Rastrigrin | 5.97E+01 | 3000 | 1 | 3.98E+01 | 3000 | 1 | 1.69E+01 | 25027 | 1 |
| Griewank | 2.95E-02 | 3000 | 1 | 0.00E+00 | 3000 | 1 | 0.00E+00 | 9055 | 1 |
| Schaffer f6 2D | 0.00E+00 | 4802 | 1 | 0.00E+00 | 3222 | 1 | 0.00E+00 | 5524 | 1 |
| Schaffer f6 | 7.82E-02 | 15223 | 2 | 7.82E-02 | 15223 | 2 | 7.82E-02 | 12138 | 2 |

**Table 3**: Results obtained from testing the BSt-PSO[c], the BSt-PSO[p], and the GP-PSO on the suite of benchmark functions showed in **Table 2**, where the set of termination conditions attained indicates which of the two sets of termination conditions was met (the constant 0.25 was used for these experiments in replacement of the constant 0.35 in the second set of termination conditions). Notice that only a single run was performed, so that the probabilistic nature of the algorithm was not considered. Thus, these are just illustrative experiments, which show that the particles find it more difficult to fine-cluster when optimizing the Rosenbrock and Schaffer f6 functions, and that it is possible that a good solution is found despite not attaining the termination conditions. The maximum number of time-steps permitted for the search to go through is 30000 in this experiment.

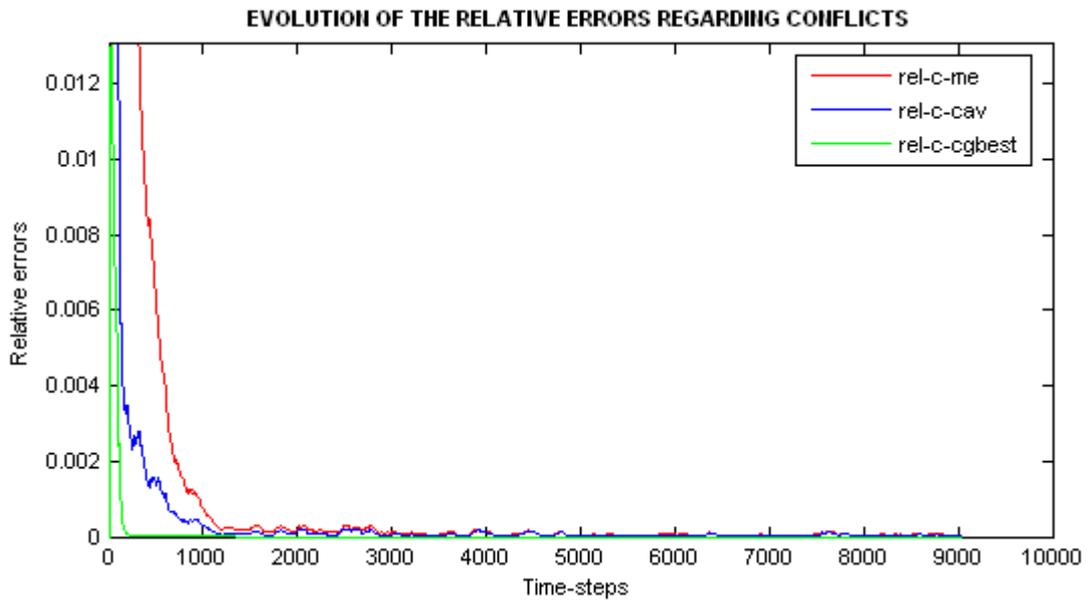

**Fig. 19**: Evolution of the relative errors regarding the conflict values for the GP-PSO optimizing the 30-dimensional Griewank function, where the feasible search-space is given by the hyper-cube $[-100,100]^2$, 30 particles are in quest for the best solution possible (minimizer), and 5 particles are in quest for the worst solution possible (maximizer). Notice that the relative errors are computed considering only the 20 particles of the minimizer whose parameters' settings favour fine-clustering.

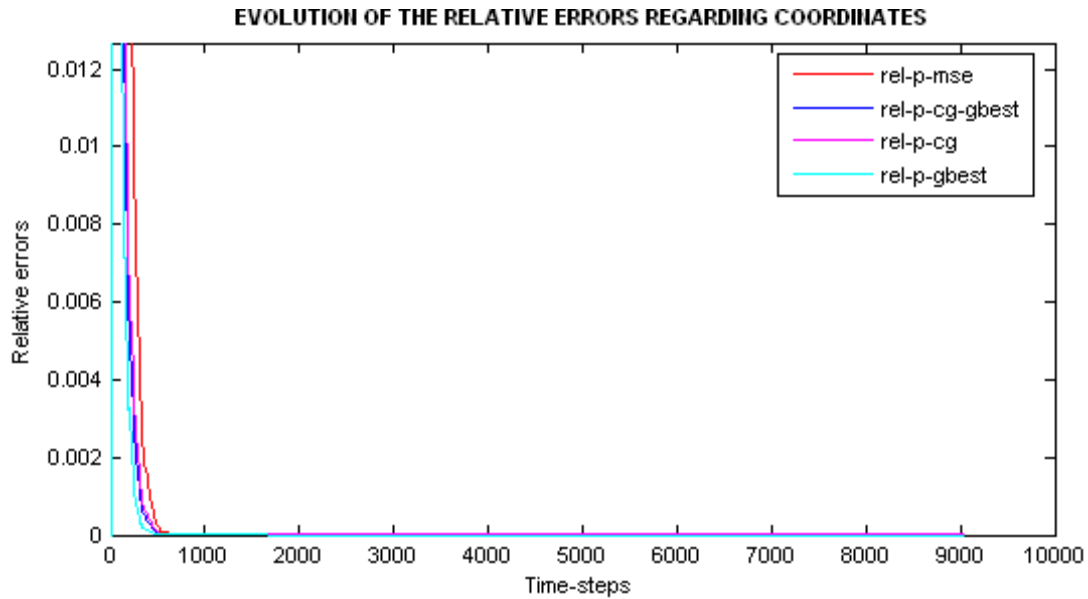

**Fig. 20**: Evolution of the relative errors regarding the particles' positions for the GP-PSO optimizing the 30-dimensional Griewank function, where the feasible search-space is given by the hyper-cube $[-100,100]^2$, 30 particles are in quest for the best solution possible (minimizer), and 5 particles are in quest for the worst solution possible (maximizer). Notice that the relative errors are computed considering only the 20 particles of the minimizer whose parameters' settings favour fine-clustering.